%% file: neurips_data_2021.tex
\documentclass{article}

    \PassOptionsToPackage{numbers, compress}{natbib}

    \usepackage[preprint]{neurips_data_2021}

\usepackage[utf8]{inputenc} 
\usepackage[T1]{fontenc}    
\usepackage{hyperref}       
\usepackage{url}            
\usepackage{booktabs}       
\usepackage{amsfonts}       
\usepackage{nicefrac}       
\usepackage{microtype}      
\usepackage{xcolor}         

\usepackage{times}
\usepackage{soul}
\usepackage[small]{caption}
\usepackage{wrapfig}
\usepackage{subfigure}
\usepackage{graphicx}
\usepackage{amsmath}
\usepackage{amsthm}
\usepackage{algorithm}
\usepackage{algpseudocode}
\algnewcommand{\LineComment}[1]{\State \(\triangleright\) #1}
\usepackage{multirow}
\usepackage{multicol}
\usepackage{rotating}
\usepackage{bbm}

\newtheorem{thm}{Theorem}

\input{math_commands.tex}

\newcommand{\xss}[1]{{#1}}



\title{A Benchmark for Low-Switching-Cost Reinforcement Learning}

\author{%
Shusheng Xu$^{1}$, Yancheng Liang$^{1}$, Yunfei Li$^{1}$, Simon Shaolei Du$^{2}$, Yi Wu$^{1,3}$\\[1ex]
	$^1$ IIIS, Tsinghua Unveristy, Beijing, \\
	$^2$ University of Washington, \\
	$^3$ Shanghai Qi Zhi institute, Shanghai  \\[1ex]
	{\tt \{xuss20, liangyc19, liyf20\}@mails.tsinghua.edu.cn}, \\
	{\tt ssdu@cs.washington.edu}, \\ {\tt jxwuyi@gmail.com}\\ 
}

\begin{document}

\maketitle

\begin{abstract}
\input{00abs.tex}

\end{abstract}

\section{Introduction}
\label{sec:intro}
\input{10introduction}

\section{Related Work}
\label{sec:rel}

\input{20rel.tex}

\section{Preliminaries}
\label{sec:pre}
\input{30pre.tex}


\section{Reinforcement Learning with Low Switching Cost}
\label{sec:feature_based_criterion}
\input{50feature_based.tex}

\section{Experiments}
\label{sec:exp}
\input{60experiment}

\section{Conclusion}
\label{sec:con}
\input{70conclusion.tex}


\bibliographystyle{plain}
\bibliography{references}

\newpage

\appendix
{\LARGE $$\makebox{\textbf{Supplementary Materials}}$$}
\section{Project Statement}

\textbf{Project website: } \emph{\footnotesize{\url{https://sites.google.com/view/low-switching-cost-rl}}}.

All of our code and detailed results can be found at our project website, including installation and running instructions for reproduciblity, \xss{and the results of individual games in Atari}. The code is hosted at GitHub under the MIT license. 

All the environments are public accessible RL testbeds. For experiments with MuJoCo engine, we register a free student license.
All experiments are run on machines with 128 CPU cores, 256G memory, and one GTX 2080 GPU card.

\label{sec: append}
\input{appendix}
\end{document}

%% file: math_commands.tex
\usepackage{amsmath,amsfonts,bm}

\def\eqref#1{equation~\ref{#1}}

\def\1{\bm{1}}

\DeclareMathAlphabet{\mathsfit}{\encodingdefault}{\sfdefault}{m}{sl}
\SetMathAlphabet{\mathsfit}{bold}{\encodingdefault}{\sfdefault}{bx}{n}

\def\sB{{\mathbb{B}}}

\DeclareMathOperator*{\argmax}{arg\,max}

\newcommand{\states}{\mathcal{S}}
\newcommand{\actions}{\mathcal{A}}

%% file: 00abs.tex
A ubiquitous requirement in many practical reinforcement learning (RL) applications, including medical treatment, recommendation system, education and robotics, is that the deployed policy that actually interacts with the environment cannot change frequently. 
Such an RL setting is called low-switching-cost RL, i.e., achieving the highest reward while reducing the number of policy switches during training.
Despite the recent trend of theoretical studies aiming to design provably efficient RL algorithms with low switching costs, none of the existing approaches have been thoroughly evaluated in popular RL testbeds. 
In this paper, we systematically studied a wide collection of policy-switching approaches, including theoretically guided criteria, policy-difference-based methods, and non-adaptive baselines.
Through extensive experiments on a medical treatment environment, the Atari games, and robotic control tasks, we present the first empirical benchmark for low-switching-cost RL and report novel findings on how to decrease the switching cost while maintain a similar sample efficiency to the case without the low-switching-cost constraint.
We hope this benchmark could serve as a starting point for developing more practically effective low-switching-cost RL algorithms. We release our code and complete results in \emph{\footnotesize{\url{https://sites.google.com/view/low-switching-cost-rl}}}.




%% file: 10introduction.tex
Reinforcement Learning (RL) has been successfully applied to solve sequential-decision problems in many real-world scenarios, such as medical domains~\cite{mahmud18applications}, robotics~\cite{gu2017deep,dmitry21mt-opt}, hardware placements~\cite{mirhoseiniZ17device,azalia20chip}, and personalized recommendation~\cite{zheng18drn}. \xss{When performing RL training in these scenarios}, it is often desirable to restrict the agent from adjusting its policy too often due to the high costs and risks of deployment. For example, \xss{updating a treatment strategy} in medical domains requires a thorough approval process by human experts~\cite{almirall12designing}; 
\xss{in personalized recommendations, it is impractical to adjust the policy online based on instantaneous data and a more common practice is to aggregate data in a period before deploying a new policy \cite{zheng18drn};
in problems where we use RL to learn hardware placements \cite{mirhoseiniZ17device}, it is desirable to limit the frequency of changes to the policy since it is costly to \xss{launch a large-scale evaluation procedure on hardware devices like FPGA}; 
for learning to control a real robot, it may be risky or inconvenient to switch the deployed policy when an operation is being performed. 
}
In these settings, it is a requirement that the deployed policy, i.e., the policy used to interact with the environment, could keep unchanged as much as possible.
Formally, we would like our RL algorithm to both produce a policy with the highest reward and at the same time reduce the number of deployed policy switches (i.e., a low \emph{switching cost}) throughout the training process.

Offline reinforcement learning~\cite{levine2020offline} is perhaps the most related framework in the existing literature that also has a capability of avoiding frequent policy deployment. 
Offline RL assumes a given transition dataset and performs RL training completely in an offline fashion without interacting with the environment at all. 
\cite{matsushima2020deployment} adopt a slightly weaker offline assumption by repeating the offline training procedure, i.e., re-collecting transition data using the current policy and re-applying offline RL training on the collected data, for about 10 times. 
However, similar to the standard offline RL methods, due to such an extreme low-deployment-constraint, the proposed solution suffers from a particularly low sample efficiency and even produces significantly lower rewards than online SAC method in many cases~\cite{matsushima2020deployment}.
In contrast with offline RL, which optimizes the reward subject to a minimal switching cost, low-switching-cost RL aims to \emph{reduce the switching cost while maintain a similar sample efficiency and final reward} compared to its unconstrained RL counterpart.


In low-switching-cost RL, the central question is \emph{how to design a criterion to decide when to update the deployed policy} based on the current training process.
Ideally, we would like this criterion to satisfy the following four properties:
\begin{enumerate}
	\item \textbf{Low switching cost:} This is the fundamental mission. An RL algorithm equipped with this policy switching criterion should have a reduced frequency to update the deployed policy. 
	\item \textbf{High reward:} Since the deployed policy can be different from the training one, the collected data can be highly off-policy.
	We need this criterion to deploy policies at the right timing so that the collected samples can be still sufficient for finally achieving the optimal reward.
	
	\item \textbf{Sample efficiency:} 
	In addition to the final reward, we also would like the algorithm equipped with such a criterion to produce a similar sample efficiency to the unconstrained RL setting without the low-switching-cost requirement.
	\item \textbf{Generality:} 
	We would like this criterion can be easily applied to a wide range of domains  rather than a specific task.
\end{enumerate}

From the theoretical side, low-switching-cost RL and its simplified bandit setting have been extensively studied~\cite{auer2002finite,cesa2013online,bai2019provably,ruan2020linear,gao2021provably, zhang2020reinforcement,zhang2020optimal}. 
The core notion in these theoretical methods is \emph{information gain}.
Specifically, they update the deployed policy only if the measurement of information gain is doubled, which also leads to optimality bounds for the final policy rewards.
We suggest the readers refer to the original papers for details of the theoretical results. We will also present algorithmic details later in Section~\ref{sec:infogain}.

However, to our knowledge, there has been no empirical study on whether these theoretically-guided criteria are in fact effective in popular RL testbeds.
In this paper, we aim to provide systematic benchmark studies on different policy switching criteria from an empirical point of view.
\xss{We remark that our experiment scenarios are based on popular deep RL environments that are much more complex than the bandit or tabular cases studied in the existing theoretical works. We hope this benchmark project could serve as a starting point towards  practical solutions that eventually solve those real-world low-switching-cost applications.}
Our contributions are summarized below.

\paragraph{Our Contributions}
\begin{itemize}
 \item We conduct the first empirical study for low-switching-cost RL on environments that require modern RL algorithms, i.e., Rainbow~\cite{hessel2018rainbow} and SAC~\cite{haarnoja2018soft}, including a medical environment, 56 Atari games\footnote{There are a total of 57 Atari games. However, only 56 of them (excluding the ``surround'' game) are supported by the atari-py package, which we adopt as our RL training interface.} and 6 MuJoCo control tasks. We test theoretically guided policy switching criteria based on the information gain as well as other adaptive and non-adaptive criteria.
\item We find that a feature-based criterion produces the best overall quantitative performance. Surprisingly, the non-adaptive switching criterion serves as a particularly strong baseline in all the scenarios and largely outperforms the theoretically guided ones.
\item Through extensive experiments, we summarize  practical suggestions for RL algorithms with with low switching cost, which will be beneficial for practitioners and future research. 
\end{itemize}

%% file: 20rel.tex
Low switching cost algorithms were first studied in the bandit setting ~\cite{auer2002finite,cesa2013online}.
Existing work on RL with low switching cost is mostly theoretical.
To our knowledge, \cite{bai2019provably} is the first work that studies this problem for the episodic finite-horizon tabular RL setting.
\cite{bai2019provably} gave a low-regret algorithm with an $O\left(H^3SA\log\left(K\right)\right)$ local switching upper bound where $S$ is the number of stats, $A$ is the number of actions, $H$ is the planning horizon and $K$ is the number of episodes the agent plays.
The upper bound was improved  in \cite{zhang2020optimal,zhang2020reinforcement}.

Offline RL (also called Batch RL) can be viewed as a close but parallel variant of low-switching-cost RL, where the policy does not interact with the environment at all and therefore does not incur any switching cost.
Offline RL methods typically learn from a given dataset~\cite{lange2012batch,levine2020offline}, and have been applied to many domains including education~\cite{mandel2014offline}, dialogue systems~\cite{jaques2019way} and robotics control~\cite{kumar2020conservative}.
Some methods interpolate offline and online methods, i.e., semi-batch RL algorithms \cite{singh1995reinforcement,lange2012batch}, which
update the policy many times on a large batch of transitions. However, reducing the switching cost during training is not their focus.
\cite{matsushima2020deployment} developed the only empirical RL method that tries to reduce the switching cost without the need of a given offline dataset.
Given a fixed number of policy deployments (i.e., 10), the proposed algorithm collects transition data using a fixed deployed policy, trains an ensemble of transition models and updated a new deployed policy via model-based RL for the next deployment iteration.
However, even though the proposed model-based RL method in \cite{matsushima2020deployment} outperforms a collection of offline RL baselines, the final rewards are still substantially lower than standard online SAC even after consuming an order of magnitude more training samples. In our work, we focus on the effectiveness of the policy switching criterion such that the overall sample efficiency and final performances can be both preserved. 
\xss{In addition to offline RL, there are also some other works that aim to reduce the interaction frequency with the environment rather than the switching cost~\cite{gu2017deep,9468918}, which is parallel to our focus. }

%% file: 30pre.tex
\textbf{Markov Decision Process: }
We consider the Markov decision model $( \mathcal{S}, \mathcal{A}, \gamma, r, p_0, P )$, where $\mathcal{S}$ is the state space, $\mathcal{A}$ is the action space, $\gamma$ is the discounted factor, $r: \states\ \times \actions \rightarrow \mathbb{R}$ is the reward function, $p_0$ is the initial state distribution, and $P(x'|x,a)$ denotes the transition probability from state $x$ to state $x'$ after taking action $a$.
A policy $\pi: \states \rightarrow \actions$ is a mapping from a state to an action, which can be either deterministic or stochastic.
An episode starts with an initial state $x_0\sim p_0$. 
At each step $h$ in this episode, the agent chooses an action $a_h$ from $\pi(x_h)$ based on the current state $x_h$, receives a reward $r(x_h,a_h)$ and moves to the next state $x_{h+1} \sim P(\cdot | x_h, a_h)$.
We assume an episode will always terminate, so each episode $e=\{(x_h^e,a_h^e)|0\le h\le H_e\}$ will always have a finite horizon $H_e$ (e.g., most practical RL environments have a maximum episode length $H_{\max}$).
The goal of the agent is to find a policy $\pi^*$ which maximizes the discounted expected reward, $\pi^\star = \arg\max_\pi \mathbb{E}_{e}\left[\sum_{h=0}^{H_e} \gamma^{h} r(x^e_h,a^e_h)\right]$.
Let $K$ denote the total transitions that the agent experienced across all the episodes during training. 
Ideally, we also want the agent to consume as few training samples as possible, i.e., a minimal $K$, to learn $\pi^\star$.
A Q-function is used to evaluate the long-term value for the action $a$ and subsequent decisions, which can be defined w.r.t. a policy $\pi$ by
\begin{equation}\label{Q-function}
Q^{\pi}(x, a):=r(x, a)+\mathbb{E}\left. \left[ \sum_{h} \gamma^h r\left(x_{h}, \pi\left(x_{h} \right)\right) \right| x_{0}=x, a_{0}=a\right].
\end{equation}


\input{33deep_rl}



%% file: 33deep_rl.tex


\textbf{Deep Off-policy Reinforcement Learning: }
	Deep Q-learning (DQN)~\cite{Mnih2015HumanlevelCT} is perhaps the most popular off-policy RL algorithm leveraging a deep neural network to approximate $Q(x, a)$. Given the current state $x_h$, the agent selects an action $a_h$ greedily based on parameterized Q-function $Q_{\theta}(x_h, a)$ and maintain all the transition data in the replay buffer.
For each training step, the temporal difference error is minimized over a batch of transitions sampled from this buffer by 
\begin{equation}\label{DQN}
\mathcal{L}(\theta)=\mathbb{E}\left[(r_{h+1} + \gamma \max_{a'} Q_{\Bar{\theta}} (x_{h+1}, a' )- Q_{\theta}(x_h, a_h))^2\right],
\end{equation}
    where $\Bar{\theta}$ represents the parameters of the target Q-network, which is periodically updated from $\theta$. 
	Rainbow~\cite{hessel2018rainbow} is perhaps the most famous DQN variant, which combines six algorithmic enhancements and achieves strong and stable performances on most Atari games.
	In this paper, we adopt a deterministic version\footnote{Standard Rainbow adds random noise to network parameters for exploration, which can be viewed as constantly switching policies over a random network ensemble. This contradicts the low-switching-cost constraint. } of Rainbow DQN as the RL algorithm for the discrete action domains. 
We also adopt count-based exploration~\cite{Tang2017ExplorationAS} as a deterministic exploration bonus. 

	 
For continuous action domains, soft actor-critic (SAC)~\cite{haarnoja2018soft} is the representative off-policy RL algorithm. SAC uses neural networks parameterized by $\theta$ to approximate both $Q(s, a)$ and the stochastic policy $\pi_{\theta}(a|s)$. $Q$-network is trained to approximate entropy-regularized expected return by minimizing 
\begin{equation}
    \mathcal{L}_{Q}(\theta) = \mathbb{E}\left[(r_{h} + \gamma(Q_{\bar{\theta}}(x_{h+1}, a') - \alpha \log \pi(a'|x_{h+1})) - Q_{\theta}(x_h, a_h))^2| a'\sim \pi(\cdot|x_{h+1})\right],
\end{equation}
where $\alpha$ is the entropy coefficient. We omit the parameterization of $\pi$ since $\pi$ is not updated w.r.t $\mathcal{L}_{Q}$. 
The policy network $\pi_\theta$ is trained to maximize $\mathcal{L}_{\pi}(\theta)=\mathbb{E}_{a\sim\pi}\left[Q(x, a) - \alpha \log \pi_{\theta}(a|x)\right]$. 

%% file: 50feature_based.tex

\begin{algorithm*}[t]
\caption{General Workflow of Low-Switching-Cost RL}
\label{alg:low-switch-rl}
    \begin{algorithmic}[1] 
        \State{Initialize parameters $\theta_{\mathrm{onl}}, \theta_{\mathrm{dep}}$, an empty replay buffer $D$, $C_{\textrm{switch}} \gets 0$} 
        \For{k $\gets$ 0 to $K-1$}
            \State{Select $a_k$ by $\pi_{\textrm{dep}}(x_k)$, execute action $a_k$  and observe reward $r_k$, state $x_{k+1}$}
            \State{Store ($x_k$, $a_k$, $r_k$, $x_{k+1}$) in $D$}
            \State{Update $\theta_{\mathrm{onl}}$ using $D$ and an off-policy RL algorithm}
            \If{$\mathcal{J}(\star) == \mathrm{true}$}
                    \State{Update $\theta_{\mathrm{dep}}\gets\theta_{\mathrm{onl}}$,\; $C_{\textrm{switch}} \gets C_{\textrm{switch}} + 1$}
            \EndIf
        \EndFor
    \end{algorithmic}
\end{algorithm*}


In standard RL, a transition $(x_h,a_h,x_{h})$ is always collected by a single policy $\pi$. 
Therefore, whenever the policy is updated, a switching cost is incurred. 
In low-switching-cost RL, we have two separate policies, a deployed policy $\pi_{\textrm{dep}}$ that interacts with the environment, and an online policy $\pi_{\textrm{onl}}$ that is trained by the underlying RL algorithm.
These policies are parameterized by $\theta_{\textrm{dep}}$ and $\theta_{\textrm{onl}}$ respectively.
Suppose that we totally collect $K$ samples during the training process, then at each transition step $k$, the agent is interacting with the environment using a deployed policy $\pi_{\textrm{dep}}^k$. After the transition is collected, the agent can decide whether to update the deployed $\pi_{\textrm{dep}}^{k+1}$ by the online policy $\pi_{\textrm{onl}}^{k+1}$, i.e., replacing $\theta_{\textrm{dep}}$ with $\theta_{\textrm{onl}}$, according to some switching criterion $\mathcal{J}$. 
Accordingly, the switching cost is defined by the  number of different deployed policies throughout the training process, namely:
\begin{align}
C_{\textrm{switch}} := \sum_{k = 1}^{K-1} \mathbb{I} \{ \pi_{\textrm{dep}}^{k-1} \not = \pi^{k}_{\textrm{dep}} \} \label{eqn:switching_cost}
\end{align}
The goal of low-switching-cost RL is to design an effective algorithm that learns $\pi^*$  using $K$ samples while produces the smallest switching cost $C_{\textrm{switch}}$. Particularly in this paper, we focus on the design of the switching criterion $\mathcal{J}$, which is the most critical component that balances the final reward and the switching cost.
The overall workflow of low-switching-cost RL is shown in Algorithm~\ref{alg:low-switch-rl}.


	In the following content, we present a collection of policy switching criteria and techniques, including those inspired by the information gain principle (Sec.~\ref{sec:infogain}) as well as other non-adaptive (Sec.~\ref{sec:linear-switch}) and adaptive criteria (Sec.~\ref{sec:policy-switch},\ref{sec:feature-switch}).
All the discussed criteria are summarized in Algorithm~\ref{alg:strategy}.
\subsection{Non-adaptive Switching Criterion}\label{sec:linear-switch}
This simplest strategy switches the deployed policy every $n$ timesteps, which we denote as ``FIX\_n'' in our experiments. 
Empirically, we notice that ``FIX\_1000'' is a surprisingly effective criteria which remains effective in most of the scenarios without fine tuning. So we primarily focus on ``FIX\_1000'' as our non-adaptive baseline.
In addition, We specifically use ``None'' to indicate the experiments without the low-switching-cost constraint where the deployed policy keeps synced with the online policy all the time. Note that this ``None'' setting is equivalent to ``FIX\_1''.

\subsection{Policy-based Switching Criterion}\label{sec:policy-switch}
Another straightforward criterion is to switch the deployed policy when the action distribution produced by the online policy significantly deviates from the deployed policy. 
Specifically, we sample a batch of training states and count the number of states where actions by the two policies differ in the discrete action domains. 
We switch the policy when the ratio of mismatched actions exceeds a threshold $\sigma_p$. For continuous actions, we use KL-divergence to measure the policy differences. 

\subsection{Feature-based Switching Criterion}\label{sec:feature-switch}
Beyond directly consider the difference of action distributions, another possible solution for measuring the divergence between two policies is through the feature representation extracted by the neural networks. 
Hence, we consider a feature-based switching criterion that decides to switch policies according to the feature similarity between different Q-networks.
Similar to the policy-based criterion, when deciding  whether to switch policy or not, 
we first sample a batch of states $\sB$ from the experience replay buffer, and then extract the features of all states with both the deployed deep Q-network and the online deep Q-network.
Particularly, we take the final hidden layer of the Q-network as the feature representation.
For a state $x$, the extracted features are denoted as $f_{\mathrm{dep}}(x)$ and $f_{\mathrm{onl}}(x)$, respectively.
The similarity score between $f_{\mathrm{dep}}$ and $f_{\mathrm{onl}}$ on state $x$ is defined as  
\begin{align}
sim(x) = \frac{\left\langle f_{\mathrm{dep}}(x) , f_{\mathrm{onl}}(x)\right\rangle}{||f_{\mathrm{dep}}(x)||\times ||f_{\mathrm{onl}}(x)||}.
\end{align}
We then compute the averaged similarity score on the batch of states $\sB$
\begin{align}
{sim(\sB)} = \frac{\sum_{x \in \sB} sim(x)}{||\sB||}.\label{eq:feature_sim}
\end{align}
With a hyper-parameter $\sigma_f \in [0,1]$, the feature-based policy switching criterion changes the deployed policy whenever $sim(\sB) \leq \sigma_f$. 

\textbf{Reset-Checking as a Practical Enhancement:}
Empirically, we also find an effective implementation enhancement, which produces lower switch costs and is more robust across different environments: we \emph{only} check the feature similarity when an episode \emph{resets} (i.e., a new episode starts) and additionally force deployment to handle extremely long episodes (e.g., in the ``Pong'' game, an episode may be trapped in loopy states and lead to an episode length of over 10000 steps).



\textbf{Hyper-parameter Selection: }
	For action-based and feature-based criteria, we uniformly sample a batch of size 512 from recent 10,000 transitions and compare the action differences or feature similarities between the deployed policy and the online policy on these sampled transitions. We also tried other sample size and sampling method, and there is no significant difference. 
	For the switching threshold (i.e., the mismatch ratio $\sigma_p$ in policy-based criterion and parameter $\sigma_f$ in feature-based criterion), we perform a rough grid search and choose the highest possible threshold that still produces a comparable final policy reward. \footnote{\xss{We list the hyper-parameter search space in Appendix~B.}}

\subsection{Switching via Information Gain}\label{sec:infogain}


Existing theoretical studies propose to switch the policy whenever the agent has gathered sufficient new information.
Intuitively, if there is not much new information, then it is not necessary to switch the policy.
To measure the sufficiency, they rely on the visitation count of each state-action pair or the determinant of the covariance matrix.
We implement these two criteria as follows.

\paragraph{Visitation-based Switching:}
Following \cite{bai2019provably}, we switch the policy when visitation count of any state-action pair reaches an exponent (specifically $2^{i}, i \in \mathbb{N}$ in our experiments). Such exponential scheme is theoretically justified with bounded switching cost in tabular cases. However, it is not feasible to maintain precise visitations for high-dimensional states, so we adopt a random projection function to map the states to discrete vectors by $\phi(x) = \textrm{sign}(A\cdot g(x))$, and then count the visitation to the hashed states as an approximation. $A$ is a fixed matrix with i.i.d entries from a unit Gaussian distribution $\mathcal{N}(0,1)$ and $g$  is a flatten function which converts $x$ to a 1-dimensional vector.

\paragraph{Information-matrix-based Switching:} Another algorithmic choice for achieving infrequent policy switches is based on the property of the feature covariance matrix~\cite{ruan2020linear,gao2021provably}, i.e., $\Lambda_h^{e} = \sum_{e:H_e\ge h}\psi(x_h^e, a_h^e)\psi(x_h^e, a_h^e)^{T} + \lambda I$, where $e$ denotes a training episode, $h$ means the $h$-th timestep within this episode, and $\psi$ denotes a mapping from the state-action space to a feature space. For each episode timestep $h$, \cite{Yasin11improved} switches the policy when the determinant of $\Lambda_h^{e}$ doubles. 
However, we empirically observe that the approximate determinant computation can be particularly inaccurate for complex RL problems. Instead, we adopt an effective alternative, i.e., switch the policy when the least absolute eigenvalue doubles.  
In practice, we again adopt a random projection function to map the state to low-dimensional vectors, $\phi(x) = \textrm{sign}(A\cdot g(x))$, and concatenate them with actions to get $\psi(x, a) = [\phi(x), a]$. 

\begin{algorithm}[tb]
\caption{Switching Criteria ($\mathcal{J}$ in Algorithm \ref{alg:low-switch-rl})}
\label{alg:strategy}
\begin{algorithmic}
\LineComment{Non-adaptive Switching}
\State{\textbf{input} environment step counter $k$, fixed switching interval $n$}
\State{\textbf{output} $bool(k\;\textrm{mod}\;n == 0)$}
\\
\LineComment{Policy-based Switching}
\State{\textbf{input} deployed and online policy $\pi_{\mathrm{dep}}, \pi_{\mathrm{onl}}$, state batch $\sB$, threshold $\sigma_p$}
\State{Compute the ratio of action difference or KL divergence for $\pi_{\mathrm{dep}}$ and $\pi_{\mathrm{onl}}$ on $\sB$ as $\delta$.}
\State{\textbf{output} $bool(\delta \geq \sigma_p)$ }
\\
\LineComment{Feature-based switching}
\State{\textbf{input} Encoder of deployed and online policy $f_{\mathrm{dep}}, f_{\mathrm{onl}}$, state batch $\sB$, threshold $\sigma_f$}
\State{Compute ${sim(\sB)}$ via Eq.(\ref{eq:feature_sim})}
\State{\textbf{output} $bool(sim(\sB) \leq \sigma_f) $ }
\\
\LineComment{Visitation-based  Switching}
\State{\textbf{input} the current visited times of state-action pair $n(\phi(x_k), a_k)$}
\State{\textbf{output} $bool(n(\phi(x_k), a_k) \in \{1, 2, 4, 8...\})$ }
\\
\LineComment{Information-matrix-based Switching}
\State{\textbf{input} episode timestep $h$, current covariance matrix $\Lambda_h^e$, old $\Lambda_h^{ \widetilde e}$ at previous switch time}
\State{Compute the least absolute eigenvalues $v_h^e$ and $v_h^{ \widetilde e}$}
\State{\textbf{output} $bool(v_h^e  \geq 2\times v_{h}^{ \widetilde e})$ }
\end{algorithmic}
\end{algorithm}

%% file: 60experiment.tex
In this section, we conduct experiments to evaluate different policy switching criteria on Rainbow DQN and SAC. For discrete action spaces, we study the Atari games and the GYMIC testbed for simulating sepsis treatment for ICU patients which requires low switching cost. For continuous control, we conduct experiments on the MuJoCo~\cite{todorov12mujoco} locomotion tasks.

\subsection{Environments}
\paragraph{GYMIC} GYMIC is an OpenAI gym environment for simulating sepsis treatment for ICU patients to an infection, where sepsis is caused by the body's response to an infection and could be life-threatening. GYMIC built an environment to simulate the MIMIC sepsis cohort, where MIMIC is an open patient EHR dataset from ICU patients. This environment generates a sparse reward, the reward is set to +15 if the patient recovers and -15 if the patient dies. This environment has 46 clinical features and a $5\times5$ action space.

\paragraph{Atari 2600} Atari 2600 games are widely employed to evaluate the performance of DQN-based agents~\cite{hessel2018rainbow}. We evaluate the efficiency of different switching criteria on a total of 56 Atari games.

\paragraph{MuJoCo control tasks} 
We evaluate different switching criteria on 6 standard continuous control benchmarks in the MuJoCo physics simulator, including Swimmer, HalfCheetah, Ant, Walker2d, Hopper and Humanoid.

\subsection{Evaluation Metric}
For GYMIC and Atari games whose action space is discrete, we adopt Rainbow DQN to train the policy; for MuJoCo tasks with continuous action spaces, we employ SAC since it is more suitable for continuous action space. We evaluate the efficiency among different switching criteria in these environments. All of the experiments are repeated over 3 seeds. Implementation details and hyper-parameters are listed in Appendix~B. All the code and the complete experiment results can be found at \emph{\footnotesize{\url{https://sites.google.com/view/low-switching-cost-rl}}}.

We evaluate different policy switching criteria based on the off-policy RL backbone and measure the reward function as well as the switching cost in both GYMIC and MuJoCo control tasks. 
For Atari games, we plot the average human normalized rewards. 
Since there are 56 Atari games evaluated, we only report the average results across all the Atari games as well as 8 representative games in the main paper. Detailed results for every single Atari game can be found at our project website.

To better quantitatively measure the effectiveness of a policy switching criterion, we propose a new evaluation metric, \emph{\textbf{Reward-threshold Switching Improvement (RSI)}}, which takes both the policy performance and the switching cost improvement into account.
Specifically, suppose the standard online RL algorithm (i.e., the ``None'' setting) can achieve an average reward of $\hat{R}$ with switching cost $\hat{C}$\footnote{We use $\hat{C}$ instead of $K$ here since some RL algorithm may not update the policy every timestep.}. Now, an low-switching-cost RL criterion $\mathcal{J}$ leads to a reward of $R_{\mathcal{J}}$ and reduced switching cost of $C_{\mathcal{J}}$ using the same amount of training samples. Then, we define RSI of criterion $\mathcal{J}$ as
\begin{equation}
\label{eq.rsi}
RSI(\mathcal{J})=\mathbbm{I}\left[R_\mathcal{J} > \left(1-\textrm{sign}(\hat{R})\sigma_{\textrm{RSI}}\right)\hat{R}\right]\log\left(\max\left(\frac{\hat{C}}{C_\mathcal{J}}, 1\right)\right),
\end{equation}
where $\mathbbm{I}[\cdot]$ is the indicator function and $\sigma_{\textrm{RSI}}$ is a reward-tolerance threshold indicating the maximum allowed performance drop with the low-switching-cost constraint applied. In our experiments, we choose a fixed threshold parameter $\sigma_{\textrm{RSI}}=0.2$. Intuitively, when the performance drop is moderate (i.e., within the threshold $\sigma_{\textrm{RSI}}$), RSI computes the logarithm \footnote{\xss{We also tried a variant of RSI that remove the log function. The results are shown in the appendix~C.}} of the relative switching cost improvements; while when the performance decreases significantly, the RSI score will be simply 0.

\begin{figure}[t]
    \centering
    \begin{minipage}[t]{0.4\linewidth}
    \centering
    \includegraphics[width=1\linewidth]{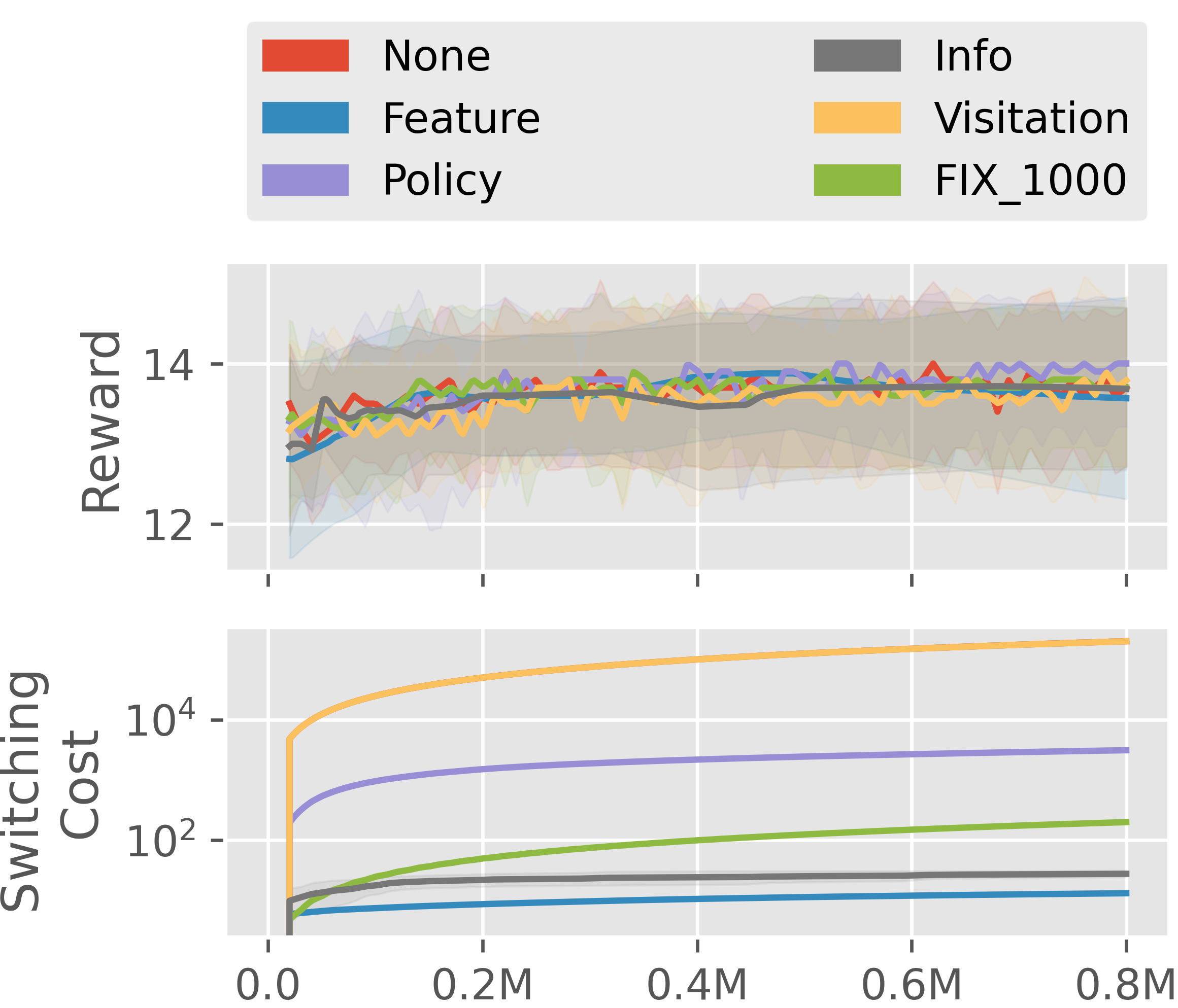}
  \caption{Results on GYMIC. 
  \textit{Top}: the learning curve of reward vs. steps. \textit{Bottom}: switching cost. Note that the switching cost of ``Visitation'' almost overlaps with ``None''.
  }
  \label{fig.gymic}
    \end{minipage}
    \hspace{2mm}
    \begin{minipage}[t]{0.53\linewidth}
    \centering
    \includegraphics[width=1\linewidth]{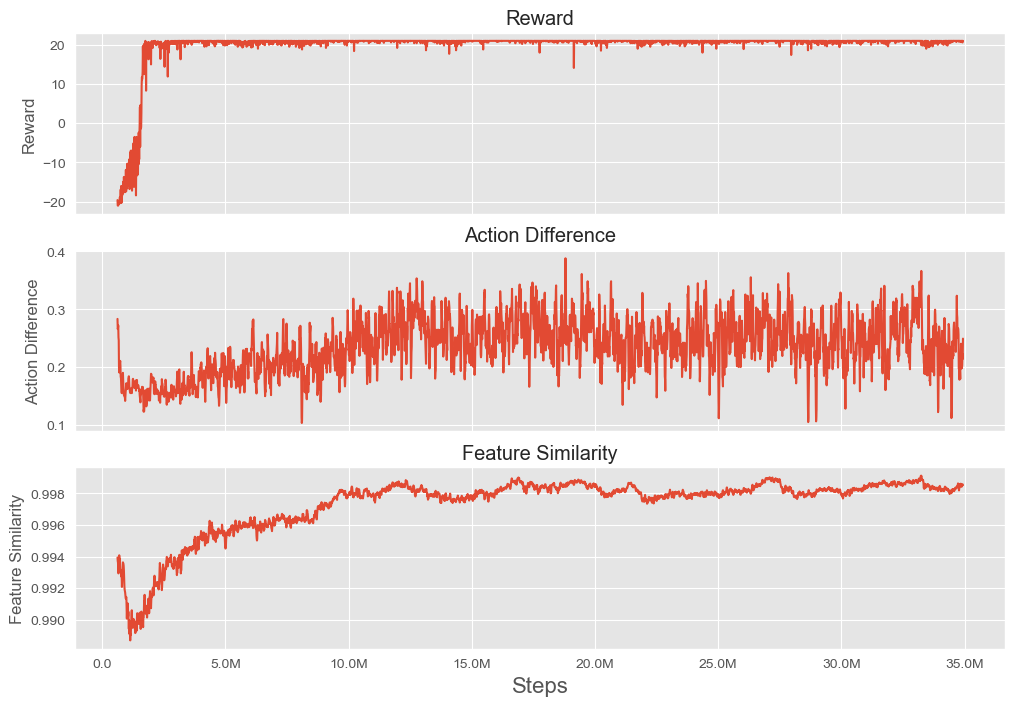}
  \caption{Action difference and feature similarity recorded on Pong. Higher feature similarity or lower action difference implies that the deployed policy and the online policy are closer.}
    \label{fig:act_d_fea_s}
    \end{minipage}
\end{figure}


\subsection{Results and Discussions}
We compare the performances of all the criteria presented in Sec.~\ref{sec:feature_based_criterion}, including unconstrained RL (``None''), non-adaptive switching (``Fix\_1000''), policy-based switching (``Policy''), feature-based switching (``Feature'') and two information-gain variants, namely visitation-based (``Visitation'') and information-matrix-based (``Info'') criteria.


\input{62results}
\label{results}

%% file: 62results.tex

\paragraph{GYMIC: }

This medical environment is relatively simple, and all the criteria achieve similar learning curves as unconstrained RL as shown in Fig.~\ref{fig.gymic}. However, the switching cost of 
visitation-based criterion is significantly higher -- 
it almost overlaps with the cost of ``None''. While the other information-gain variant, i.e., information-matrix-based criterion, performs much better in this scenario.
Overall, feature-based criterion produces the most satisfactory switching cost without hurt to sample efficiency.
\xss{Note that in GYMIC, different methods all have very similar performances and induce a particularly high variance across seeds. This is largely due to the fact the intrinsic transition dynamics in GYMIC is highly stochastic.  
GYMIC simulates the process of treating ICU patients. Each episode starts from a random state of a random patient, which varies a lot during the whole training process. In addition, GYMIC only provides a binary reward (+15 or -15) at the end of each episode for whether the patient is cured or not, which amplifies the variance. 
Furthermore, we notice that even a random policy can achieve an average reward of 13, which is very close to the maximum possible reward 15. Therefore, possible improvement by RL policies can be small in this case.
However, we want to emphasize that despite similar reward performances, different switching criteria lead to significantly different switching costs.
}

\begin{figure}
\begin{center}
    \includegraphics[width=0.8\linewidth]{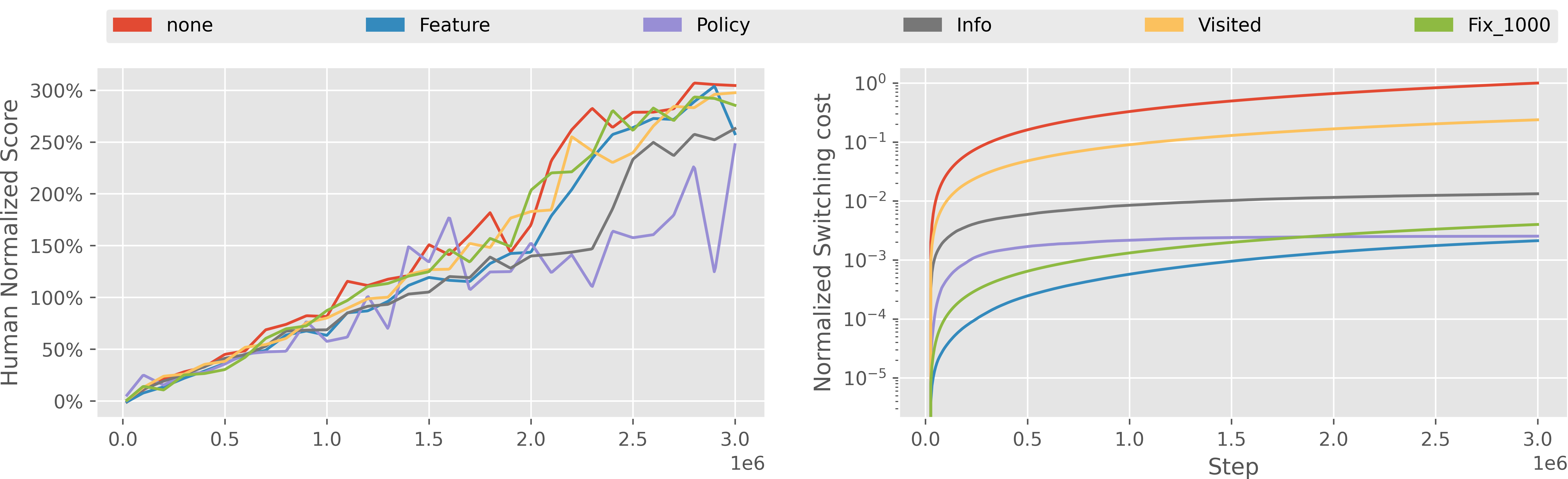}
\end{center}
\caption{The average results on Atari games. We compare different switching criteria across 56 Atari games with 3 million training steps. We visualize the \emph{human normalized reward} on the left. The figure on the right shows the average switching cost, which is normalized by the switching cost of ``None'' and shown in a log scale.}
\label{fig.atari_average}
\end{figure}

\begin{figure}[ht]
\centering
    \includegraphics[width=0.8\linewidth]{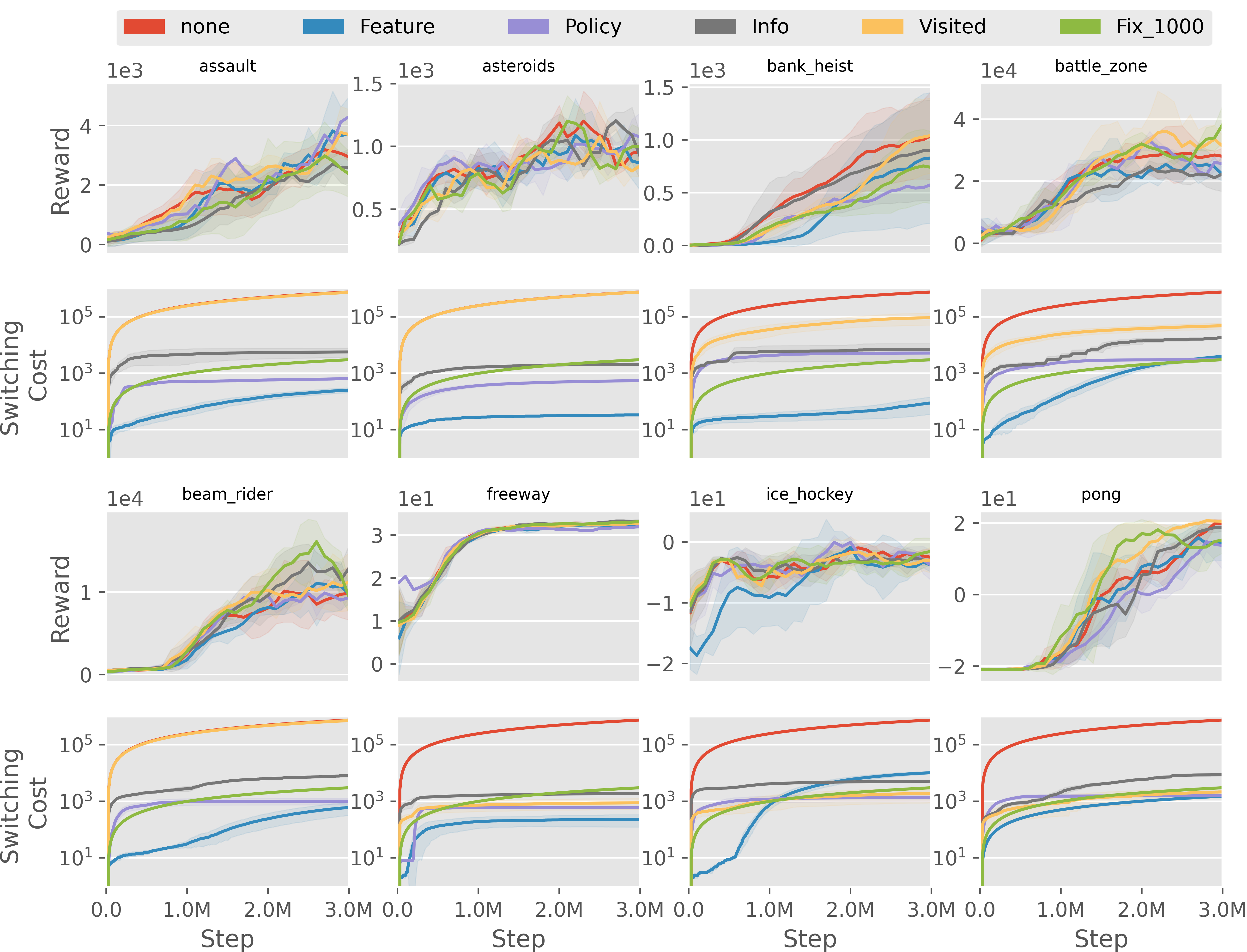}
\caption{The results on several representative Atari games. In each environment, we visualize the training reward over the steps on the top and the switching cost in a $\log$ scale at the bottom. 
}
\label{fig.atari}
\vspace{-4mm}
\end{figure}

\paragraph{Atari Games: }
We then compare the performances of different switching criteria in the more complex Atari games. The state spaces in Atari games are images, which are more complicated than the low-dimensional states in GYMIC. Fig.~\ref{fig.atari_average}  shows the average reward and switching of different switching criteria across all the 56 games, where the feature-based solution leads to the best empirical performance. We also remark that the non-adaptive baseline is particularly competitive in Atari games and outperforms all other adaptive solutions except the feature-based one.
We also show the results in 8 representative games in Fig.~\ref{fig.atari}, including the reward curves (odd rows) and switching cost curves (even rows). We can observe that information-gain variants produce substantially more policy updates while the feature-based and non-adaptive solutions are more stable.

In addition, we also noticed that the policy-based solution is particularly sensitive to its hyper-parameter in order to produce desirable policy reward, which suggests that the neural network features may change much more smoothly than the output action distribution.

To validate this hypothesis, we visualize the action difference and feature difference of the unconstrained Rainbow DQN on the Atari game ``Pong'' throughout the training process in Fig.~\ref{fig:act_d_fea_s}. Note that in this case, the deployed policy is synced with the online policy in every training iteration, so the difference is merely due to a single training update. However, even in a unconstrained setting, the difference of action distribution fluctuates significantly. By contrast, the feature change is much more stable. 
We also provide some theoretical discussions on feature-based criterion in Appendix~D. 


\begin{figure}[t]
\centering
    \includegraphics[width=0.8\linewidth]{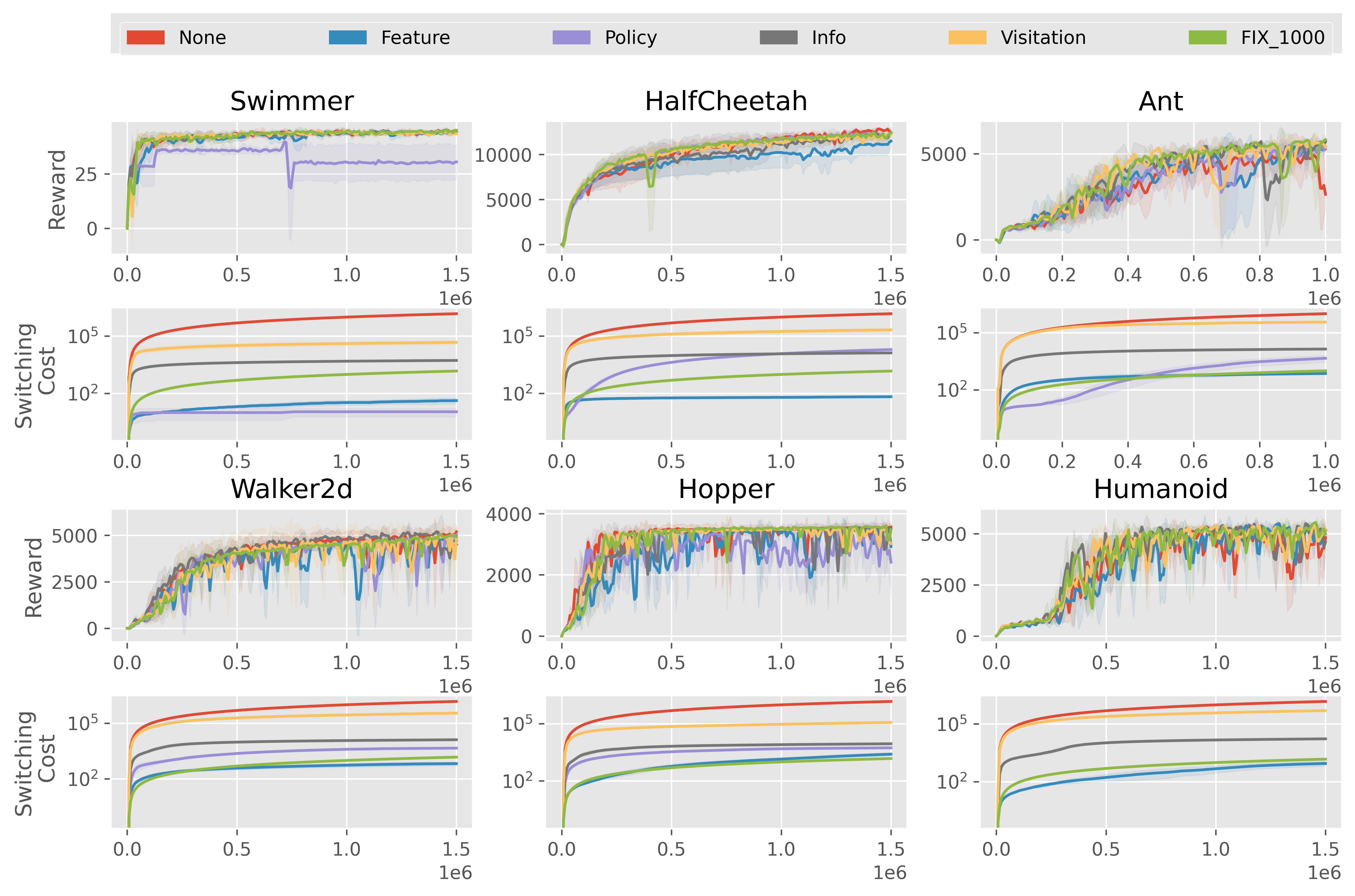}
\caption{The results on MuJoCo tasks. 
}
\label{fig.mujoco}
\vspace{-3mm}
\end{figure}

\paragraph{MuJoCo Control:} 
We evaluate the effectiveness of different switching criteria with SAC on all the 6 MuJoCo continuous control tasks. The results are shown in Fig.~\ref{fig.mujoco}.
In general, we can still observe that the feature-based solution achieves the lowest switching cost among all the baseline methods while the policy-based solution produces the most unstable training. Interestingly, although the non-adaptive baseline has a relatively high switching cost than the feature-based one, the training curve has the less training fluctuations, which also suggests a future research direction on incorporating training stability into the switching criterion design. 


\begin{table}[ht]
    \caption{RSI (Eq.~\ref{eq.rsi}, $\sigma=0.2$) for different criteria over different domains. We take unconstrained RL (i.e., ``None'') performance as the RSI reference, so the RSI value for ``None'' is always zero. }
    \label{tab.rsi}
    \centering
    \begin{tabular}{cccccc}
    \toprule
       Avg. RSI & Feature & Policy & Info & Visitation & FIX\_1000 \\
    \midrule
        GYMIC & \textbf{9.63} & 4.16 &  8.88 & 0.0 &  6.91 \\
        Atari & \textbf{3.61} & 2.82 & 2.11 & 1.81 & 3.15 \\
        Mujoco & \textbf{8.20} & 3.45 & 4.83 & 1.92 & 6.91 \\
    \bottomrule
    \end{tabular}
\end{table}

\textbf{Average RSI Scores:} Finally, we also report the RSI scores of different policy switching criteria on different domains. For each domain, we compute the average value of RSI scores over each individual task in this domain. The results are reported in Table~\ref{tab.rsi}, where we can observe that the feature-based method consistently produces the best quantitative performance across all the 3 domains. \footnote{\xss{We list the results of $\sigma_{RSI}=0.2$ in Talbe~\ref{tab.rsi}, and also evaluate RSI using other $\sigma_{RSI}$. See appendix~C for details.}} 

\xss{In addition, we apply a $t$-test for the switching costs among different criteria. The testing procedure is carried out for each pair of methods over all the experiment trials across all the environments and seeds. For Atari and MuJoCo, the results show that there are significant differences in switching cost between any two criteria (with $p$-value < 0.05). For GYMIC, ``None’’ and ``Visited’’ is the only pair with no significant difference. It is also worth mentioning that RSI for ``Visitation’’ for GYMIC is 0, which also shows the switching costs of ``Visitation’’ and ``None’’ are nearly the same.}

%% file: 70conclusion.tex
In this paper, we focus on low-switching-cost reinforcement learning problems and take the first empirical step towards designing an effective solution for reducing the switching cost while maintaining good performance.
By systematic empirical studies on practical benchmark environments with modern RL algorithms, we find the existence of a theory-practice gap in policy switching criteria and suggest a feature-based solution can be preferred in practical scenarios.

\xss{
We remark that most theoretical works focus on simplified bandits or tabular MDPs when analysing mathematical properties of information-gain-based methods, but these theoretical insights are carried out on much simplified scenarios compared with popular deep RL testbeds that we consider here.
In addition, these theoretical analyses often ignore the constant factors, which can be crucial for the practical use. By contrast, feature-based methods are much harder to analyze theoretically particularly in the use of non-linear approximators, which, however, produces much better practical performances.
This also raises a new research direction for theoretical studies.
}
\xss{We emphasize that our paper not only provides a benchmark but also raises many interesting open questions for algorithm development.}
For example, although feature-based solution achieves the best overall performance, it does not substantially outperform the the naive non-adaptive baseline. It still has a great research room towards designing a more principled switching criteria.
Another direction is to give provable guarantees for these policy switching criteria that work for methods dealing with large state space in contrast to existing analyses about tabular RL~\cite{bai2019provably,zhang2020optimal,zhang2020reinforcement}.
We believe our paper is just the first step on this important problem, which could serve as a foundation towards great future research advances.

Thanks to the strong research nature of this work, we believe our paper does not produce any negative societal impact. 



%% file: appendix.tex
\section{Experiment Details}
\label{apd:detail}
\subsection{Environments}
\paragraph{GYMIC} GYMIC is an OpenAI gym environment for simulating sepsis treatment for ICU patients to an infection, where sepsis is caused by the body's response to an infection and could be life-threatening. GYMIC built an environment to simulate the MIMIC sepsis cohort, where MIMIC is an open patient EHR dataset from ICU patients. This environment generates a sparse reward, the reward is set to +15 if the patient recovered and -15 if the patient died. This environment has 46 clinical features and a $5\times5$ action space.

\paragraph{Atari 2600} Atari 2600 games are widely employed to evaluate the performance of DQN based agents. We evaluate the efficiency among different switching criteria on 56 games. 

\paragraph{MuJoCo} 
MuJoCo contains continuous control tasks running in a physics simulator, we evaluate different switching criteria on 6 locomotion benchmarks.

For GYMIC and Atari games whose action space is discrete, we adopt Rainbow DQN to train the policy, for MuJoCo tasks with continuous action spaces, we employ SAC since it is more suitable for continuous action space.

\subsection{Hyper-parameters of deterministic Rainbow}
Table \ref{tab:rainbow-hyper-parameters} lists the basic hyper-parameters of Rainbow. All of our experiments share these hyper-parameters except that the experiments on GYMIC adopt $H_{\mathrm{target}} = 1K$. Most of these hyper-parameters are the same as those in the original Rainbow algorithm. For count-based exploration, the bonus $\beta$ is set to 0.01.
\begin{table}[ht]
    \caption{The basic hyper-parameters of Rainbow. we used the Adam optimizer with learning rate $\alpha=0.0000625$ and $\epsilon=1.5\times10^{-4}$
    . Before training the online policy, we let the initialized random policy run 20K steps to collect some transitions. The capacity of the replay buffer is 1M.
    During the training process, we sample 32 transitions from the replay buffer and update the online policy every four steps.
    The reward is clipped into [-1, 1] and ReLU is adopted as the activation function. For replay prioritization we use the recommended
    proportional variant, with importance sampling from 0.4 to 1, the prioritization $\omega$ is set to 0.5.
    In addition, we employ $N_{\mathrm{atoms}} = 51, V_{\mathrm{min}} = -10, V_{\mathrm{max}} = 10$ for distributional RL and $n=3$ for multi-step returns. The count-based bonus is set to 0.01.}
    \label{tab:rainbow-hyper-parameters}
    \centering
    \begin{tabular}{l|c}
        \hline
        Parameter & Value  \\
        \hline
        $H_{\mathrm{start}}$ & 20K\\
        learning rate & 0.0000625 \\
        $H_{\mathrm{target}}$(Atari) & 8K \\
        $H_{\mathrm{target}}$(GYMIC) & 1K \\
        Adam $\epsilon$ & $1.5\times 10^{-4}$ \\
        Prioritization type & proportional \\
        Prioritization exponent $\omega$ & 0.5 \\
        Prioritization importance sampling & $ 0.4 \xrightarrow{} 1.0$\\
        Multi-step returns n & 3 \\
        Distributional atoms $N_{\mathrm{atoms}}$ & 51 \\
        Distributional $V_{\mathrm{min}}, V_{\mathrm{max}}$ & [-10, 10] \\
        Discount factor $\gamma$ & 0.99 \\
        Memory capacity N & 1M \\
        Replay period & 4 \\
        Minibatch size & 32 \\ 
        Reward clipping & [-1, 1] \\
        Count-base bonus & 0.01 \\
        Activation function $\beta$ & ReLU \\
        \hline
        
    \end{tabular}
\end{table}

Table \ref{tab:sepsis-hyper-parameters} lists the extra hyper-parameters for experiments on GYMIC. Since there are 46 clinical features in this environment, we stack 4 consecutive states to compose a 184-dimensional vector as the input for the state encoder. The state encoder is a 2-layer MLP with hidden size 128.
\begin{table}[ht]
    \caption{Extra hype-parameters for the experiments in GYMIC, we stack 4 consecutive states and adopt a 2-layer MLP with hidden size 128
    to extract the feature of states.}
    \label{tab:sepsis-hyper-parameters}
    \centering
    \begin{tabular}{l|c}
        \hline
        Parameter & Value  \\
        \hline
        State Stacked & 4 \\
        Number of layers for MLP & 2\\
        Hidden size & 128 \\
        \hline
    \end{tabular}
\end{table}

Table \ref{tab:atari-hyper-parameters} shows the additional hyper-parameters for experiments on Atari games. The observations are grey-scaled and resized to tensors with size $84 \times 84$, and 4 consecutive frames are concatenated as a single state. Each action selected by the agent is repeated for 4 times. The state encoder is composed of 3 convolutional layers with 32, 64 and 64 channels, which use 8x8, 4x4, 3x3 filters and strides of 4, 2, 1, respectively.
\begin{table}[ht]
    \caption{Additional hyper-parameters for experiments in Atari games. Observations are grey-scaled and rescaled to $84\times 84$.
    4 consecutive frames are staked as the state and each action is acted for four times. We limit the max number of frames in an
    episode to 108K. The state encoder consists of 3 convolutional layers.}
    \label{tab:atari-hyper-parameters}
    \centering
    \begin{tabular}{l|c}
        \hline
        Parameter & Value  \\
        \hline
        Gray scaling & True\\
        Observation & (84, 84) \\
        Frame Stacked & 4 \\
        Action repetitions & 4 \\
        Max frames per episode & 108k \\
        Encoder channels & 32, 64, 64 \\
        Encoder filter size & $8\times8, 4\times4, 3\times3$ \\
        Encoder stride & 4, 2, 1 \\
        \hline
    \end{tabular}
\end{table}

\subsection{Hyper-parameters of SAC}
We list the hyper-parameters of SAC in Table~\ref{tab:SAC-hyper-parameters}. We adopt Adam as the optimizer, the learning rate is set to 0.001 for MuJoCo control tasks except for Swimmer which is 0.0003, and the temperature parameter $\alpha$ is set to 0.2 for Mujoco control tasks except for Humanoid which is 0.05. Other hyper-parameters are the same for these tasks. The update frequency ``50/50'' means we perform 50 iterations to update the online policy per 50 environment steps.
\begin{table}[hb]
    \caption{Hyper-parameters of SAC algorithms on MuJoCo control tasks. 
    }
    \label{tab:SAC-hyper-parameters}
    \centering
    \begin{tabular}{l|c}
        \hline
        Parameter & Value  \\
        \hline
        Warm-up samples before training starts & 10K\\
        Learning rate & 0.001 (0.0003 for Swimmer) \\
        Optimizer & Adam \\ 
        Temperature parameter $\alpha$ & 0.2 (0.05 for Humanoid)  \\
        Discount factor $\gamma$ & 0.99 \\
        Memory capacity N & 1M \\
        Number of hidden layers & 2 \\
        Number of hidden nodes per layer & 256 \\
        Target smoothing coefficient & 0.005 \\
        Update frequency & 50/50 \\
        Target update interval & 1 \\
        Minibatch size & 128 \\ 
        Activation function  & ReLU \\
        \hline
    \end{tabular}
\end{table}

\subsection{Hyper-parameters of different criteria}
\xss{
For the switching threshold in policy-based and feature-based criteria (i.e., the mismatch ratio $\sigma_p$ in policy-based criterion and parameter $\sigma_f$ in feature-based criterion), we perform a rough grid search and choose the highest possible threshold that still produces a comparable final policy reward. For GYMIC, we tried $\sigma_p \in \{0.25, 0.5\}$ and $\sigma_f \in \{0.97, 0.98, 0.99\}$ and finally adopted $\sigma_p = 0.5$ and $\sigma_f = 0.97$. For Atari games, we tried $\sigma_p \in \{0.25, 0.5\}$ and $\sigma_f \in \{0.98, 0.99\}$. For MuJoCo, we tried $\sigma_p \in \{0.5, 1.0, 1.5\}$\footnote{\xss{$\{0.5, 1.0, 1.5\}$ is the thresholds of KL.}} and $\sigma_f \in \{0.7, 0.8, 0.9\}$
}
\section{Discussion on RSI}
\subsection{RSI without log function}
\xss{
We choose log function because we observe that switching costs of different criteria vary in orders of magnitudes.
We also tried a variant of RSI that removes the log function, which is
\begin{equation}
\mathbbm{I}\left[R_\mathcal{J} > \left(1-\textrm{sign}(\hat{R})\sigma_{\textrm{RSI}}\right)\hat{R}\right]\left(\max\left(\frac{\hat{C}}{C_\mathcal{J}}, 1\right)\right),
\end{equation}
The results are shown in the following Table~\ref{tab:rsi_variant} ($\sigma_{RSI}$ = 0.2 ). We think these numbers have exaggerated the differences among different criteria and we finally adopt the log function. 
}
\begin{table}[ht]
    \caption{\xss{The results of RSI variant that removes the log function, the results exaggerate the differences among different criteria.}}
    \label{tab:rsi_variant}
    \centering
    \begin{tabular}{c|ccccc}
    \hline
    & Feature & Policy & Info & Visitation & Fix\_1000 \\
    \hline
    MYGIC & \textbf{15152} & 64 & 7210 & 1 & 1000 \\
    Atari & \textbf{1214} & 154 & 98 & 76 & 571 \\
    MuJoCo & \textbf{10516} & 133 & 139 & 10 & 1000 \\
    \hline
    \end{tabular}
\end{table}

\subsection{Evaluate RSI with different $\sigma_{RSI}$}
\xss{
we evaluate RSI using different $\sigma_{RSI}$, and list the results in Table~\ref{tab:different_sigma}. Since the rewards obtained by different criteria on GYMIC are almost the same, RSI remains mostly unchanged when using different $\sigma_{RSI}$. The only exception is when $\sigma_{RSI} = 0$. In this case, RSI of ``Feature’’ and ``Info'' become 0. This is because a positive RSI requires strictly better or equal reward performance than ``None’’ criterion when $\sigma_{RSI}$ = 0. For Atari and MuJoCo, the overall trend is that a larger $\sigma_{RSI}$ results in a larger RSI, since a larger $\sigma_{RSI}$ tolerates a wider performance range. And we can observe that for most $\sigma_{RSI}$, the conclusions remain unchanged.
}
\begin{table}[ht]
    \caption{\xss{The RSI of different $\sigma_{RSI}$ }}
    \label{tab:different_sigma}
    \centering
    \begin{tabular}{c|cccccccccc}
    \toprule
    \multicolumn{11}{c}{MYGIC} \\
    \midrule
        $\sigma_{RSI}$ & 0.0 & 0.1 & 0.2 & 0.3 & 0.4 & 0.5 & 0.6 & 0.7 & 0.8 & 0.9 \\
    \midrule
        Feature & 0.0 & \textbf{9.63} & \textbf{9.63} & \textbf{9.63} & \textbf{9.63} & \textbf{9.63} & \textbf{9.63} & \textbf{9.63} & \textbf{9.63} & \textbf{9.63} \\ 
        Policy & 4.16 & 4.16 & 4.16 & 4.16 & 4.16 & 4.16 & 4.16 & 4.16 & 4.16 & 4.16 \\
        Info & 0.0 & 8.88 & 8.88 & 8.88 & 8.88 & 8.88 & 8.88 & 8.88 & 8.88 & 8.88 \\
        Visitation & 0.0 & 0.0 & 0.0 & 0.0 & 0.0 & 0.0 & 0.0 & 0.0 & 0.0 & 0.0\\
        FIX\_1000 & \textbf{6.91} & 6.91 & 6.91 & 6.91 & 6.91 & 6.91 & 6.91 & 6.91 & 6.91 & 6.91 \\
    \toprule
    \multicolumn{11}{c}{Atari} \\
    \midrule
        $\sigma_{RSI}$ & 0.0 & 0.1 & 0.2 & 0.3 & 0.4 & 0.5 & 0.6 & 0.7 & 0.8 & 0.9 \\
    \midrule
        Feature & 1.66 & \textbf{2.83} & \textbf{3.61} & \textbf{3.85} & 4.03 & \textbf{4.66} & \textbf{5.0} & \textbf{5.2} & \textbf{5.49} & \textbf{5.49} \\
        Policy & 1.56 & 2.22 & 2.82 & 3.56 & 4.04 & 4.39 & 4.61 & 4.78 & 4.78 & 4.78 \\
        Info & 0.81 & 1.46 & 2.11 & 2.89 & 3.23 & 3.66 & 4.19 & 4.27 & 4.35 & 4.35 \\
        Visitation & 1.09 & 1.48 & 1.81 & 1.96 & 2.19 & 2.21 & 2.21 & 2.28 & 2.28 & 2.28 \\
        FIX\_1000 & \textbf{2.27} & 2.76 & 3.15 & 3.84 & \textbf{4.34} & 4.63 & 4.93 & 5.03 & 5.03 & 5.03 \\
    \toprule
    \multicolumn{11}{c}{MuJoCO} \\
    \midrule
        $\sigma_{RSI}$ & 0.0 & 0.1 & 0.2 & 0.3 & 0.4 & 0.5 & 0.6 & 0.7 & 0.8 & 0.9 \\
    \midrule
        Feature & 2.44 & 6.54 & \textbf{8.20} & \textbf{8.20} & \textbf{8.20} & \textbf{8.20} & \textbf{8.20} & \textbf{8.20} & \textbf{8.20} & \textbf{8.20} \\
        Policy & 1.76 & 3.45 & 3.45 & 4.38 & 6.35 & 6.35 & 6.35 & 6.35 & 6.35 & 6.35 \\
        Info & 2.40 & 4.83 & 4.83 & 4.83 & 4.83 & 4.83 &4.83 & 4.83 & 4.83 & 4.83 \\
        Visitation & 0.35 & 1.92 & 1.92 & 1.92 & 1.92 & 1.92 & 1.92 & 1.92 & 1.92 & 1.92 \\
        FIX\_1000 & \textbf{3.45} & \textbf{6.91} & 6.91 & 6.91 & 6.91 & 6.91 & 6.91 & 6.91 & 6.91 & 6.91\\
    \bottomrule
    \end{tabular}
\end{table}

\section{Discussion on Feature-based Criterion}\label{sec:theorem}
The feature-based criterion can be intuitively justified using some recent advances in representation learning theory.
To illustrate the insight, we consider the following setting. 
Suppose we want to learn $f(\cdot)$, a representation function that maps the input to a $k$-dimension vector.
We assume we have input-output pairs $(x,y)$ with $y = \left\langle w, f^*(x)\right\rangle
$ for some underlying representation function $f^*(\cdot)$ and a linear predictor $w \in \mathbb{R}^k$.
For ease of presentation, let us assume we know $w$, and our goal is to learn the underlying representation which together with $w$ gives us $0$ prediction error.

Suppose we have data sets $\mathcal{D}_1$ and $\mathcal{D}_2$. We use $\mathcal{D}_1$ to train an estimator of $f^*$, denoted as $f^1$, and $\mathcal{D}_1 \cup \mathcal{D}_2$ to train another estimator of $f^*$, denoted as $f^{1+2}.$
The training method is empirical risk minimization, i.e.,
\begin{align*}
f^1 \leftarrow \min_{f \in \mathcal{F}} \frac{1}{\left|\mathcal{D}_1\right|}\sum_{(x,y) \in \mathcal{D}_1}\left(y-\left\langle w, f(x) \right\rangle\right)^2\text{ and } \\
f^{1+2} \leftarrow \min_{f \in \mathcal{F}} \frac{1}{\left|\mathcal{D}_{1}\cup \mathcal{D}_2\right|}\sum_{(x,y) \in \mathcal{D}_{1}\cup \mathcal{D}_2}\left(y-\left\langle w, f(x) \right\rangle\right)^2
\end{align*}
where $\mathcal{F}$ is some pre-specified representation function class.

The following theorem suggests if the similarity score between $f^1$ and $f^{1+2}$ is small, then $f^{1}$ is also far from the underlying representation $f^*$.

\begin{thm}
	\label{thm:feature}
Suppose $f^1$ and $f^{1+2}$ are trained via aforementioned scheme.
There exist dataset $\mathcal{D}_1$, $\mathcal{D}_2$, function class $\mathcal{F}$ and $w$ such that if the similarity score between $f^1$ and $f^{1+2}$ on $\mathcal{D}_{1+2}$ is smaller than $\alpha$, then the prediction error of $f^1$ on $\mathcal{D}_{1+2}$ is $1-\alpha$.
\end{thm}

Theorem~\ref{thm:feature} suggests that in certain scenarios, if the learned representation has not converged (the similarity score is small), then it cannot be the optimal representation which in turn will hurt the prediction accuracy.
Therefore, if the similarity score is small, we should change the deployed policy.
\label{sec:proof}

\begin{proof}[Proof of Theorem~1]
We let $w=\left(1,1,\ldots,1\right) \in \mathbb{R}^k$ be a $k$-dimensional all one vector.
We let 
\begin{align*}
    \mathcal{F} = \{ f: f(x) = (& 2\sigma(\left\langle v_1, x\right\rangle)-1,2\sigma(\left\langle v_2, x\right\rangle)-1,\ldots,\\
    & 2\sigma(\left\langle v_k, x\right\rangle)-1)  \} \subset \{\mathbb{R}^k \rightarrow \mathbb{R}^k\}&&
\end{align*}
with $\sigma(\cdot)$ being the ReLU activation function\footnote{We define $\sigma(0)=0.5$} and $v_i \in \{e_i,-e_i\}$ where $e_i \in \mathbb{R}^k$ denotes the vector that only the $i$-th coordinate is $1$ and others are $0$.
We assume $k$ is an even number and $\alpha k$ is an integer for simplicity.
We let the underlying $f^*$ be the vector correspond to $\left(e_1,e_2,\ldots,e_k\right)$.
We let $\mathcal{D}_1 = \{(e_1,1),(e_2,1),\ldots,(e_{(1-\alpha)k},1)\}$ and $\mathcal{D}_2=\{(e_{(1-\alpha)k+1},1),\ldots,(e_{k},1)\}$.
Because we use the ERM training scheme, it is clear that the training on $\mathcal{D}_1 \cup \mathcal{D}_2$ will recover $f^*$, i.e., $f^{1+2} = f^*$ because if it is not $f^*$ is better solution ($f^*$ has $0$ error ) for the empirical risk.
 Now if the similarity score between $f^1$ and $f^{1+2}$ is smaller  than $\alpha$, it means for $f^1$, its corresponding $\{v_{(1-\alpha)k+1},\ldots,v_{k}\}$ are not correct.
 In this case, $f^1$'s prediction error is at least $1-\alpha$ on $\mathcal{D}_1 \cup \mathcal{D}_2$, because it will predict $0$ on all inputs of $\mathcal{D}_2$.
 
\end{proof}